\documentclass[letterpaper, 10pt, conference,twocolumn]{article}  
                                                         
\usepackage[margin=0.6in]{geometry}

\usepackage[pagestyles]{titlesec}
\usepackage[titles,subfigure]{tocloft}
\titleformat*{\section}{\Large\bfseries}
\titleformat*{\subsection}{\large\bfseries}
\titleformat*{\subsubsection}{\normalsize\bfseries}
\usepackage{graphics} 
\usepackage{xcolor}
\usepackage[colorlinks=true,citecolor=blue,linkcolor=blue,citecolor=blue,allcolors=blue]{hyperref}%
\date{\vspace{-5ex}}
\usepackage{amsmath} 
\usepackage{amssymb}  
\usepackage[sort&compress,numbers]{natbib}
\usepackage{authblk}
\usepackage[heightadjust=all]{floatrow}
\usepackage{amsmath,amssymb,amsfonts}
\usepackage{algorithmic}
\usepackage{subcaption}
\usepackage{graphicx}
\usepackage{textcomp}
\usepackage{makecell}
\usepackage{xcolor}
\def\BibTeX{{\rm B\kern-.05em{\sc i\kern-.025em b}\kern-.08em
    T\kern-.1667em\lower.7ex\hbox{E}\kern-.125emX}}
\usepackage{hyperref}
\usepackage[utf8]{inputenc}
\usepackage{dsfont}
\usepackage{bm}
\usepackage{mathrsfs}
\usepackage{tikz}
\usepackage{graphicx}
\usepackage{algorithmic}
\usepackage{algorithm}
\usepackage{paralist}
\usepackage{listings,xspace}
\usepackage{soul}
\usepackage{latexsym}
\usepackage{graphicx,psfrag}
\usepackage{amsmath,amssymb,amsthm,mathtools}
\usepackage{color,cite}

\title{\LARGE \bf \alg: A Physics-Constrained Deep Continuous\\Space-Time Super-Resolution Framework}

\author[1]{Chiyu\,“Max”\,Jiang{\footnote{Denotes equal contribution.}}}
\author[2]{Soheil\,Esmaeilzadeh$^*$}
\author[3]{Kamyar\,Azizzadenesheli}
\author[4]{Karthik\,Kashinath}
\author[4]{Mustafa\,Mustafa}
\author[2]{Hamdi\,A.\,Tchelepi}
\author[1]{Philip\,Marcus}
\author[4]{Prabhat}
\author[3,5]{Anima\,Anandkumar}
\affil[1]{\footnotesize University of California, Berkeley, CA 94720, USA}
\affil[2]{\footnotesize Stanford University, Stanford, CA 94305, USA}
\affil[3]{\footnotesize California Institute of Technology, Pasadena, CA, 91125, USA}
\affil[4]{\footnotesize Lawrence Berkeley National Laboratory, Berkeley, CA 94720, USA}
\affil[5]{\footnotesize NVIDIA, Santa Clara, CA 95051, USA}

\newcommand{\Domain}{\Omega}
\newcommand{\DomainP}{\Omega_{Spatial}}
\newcommand{\D}{\mathcal{D}}
\newcommand{\C}{\mathcal{C}}
\newcommand{\G}{\mathcal{G}}

\newcommand{\F}{\mathcal{F}}

\newcommand{\Real}{\mathbb{R}}

\definecolor{dukeblue}{rgb}{0.0, 0.0, 0.61}
\newcommand{\bb}[1]{\textbf{\textcolor{dukeblue}{#1}}}

\newcommand{\Operator}{\Gamma}
\newcommand{\BOperator}{\Lambda}
\newcommand{\GOperator}{{\mathcal{H}}}
\newcommand{\LOperator}{{\mathcal{L}}}

\newcommand{\alg}{\textsc{MeshfreeFlowNet}\xspace}

\newcommand{\PDE}{\textsc{PDE}\xspace}

\begin{document}

\maketitle
\thispagestyle{empty}
\pagestyle{empty}

\begin{abstract}
We propose \alg, a novel deep learning-based super-resolution framework to generate continuous (grid-free) spatio-temporal solutions from the low-resolution inputs. While being computationally efficient, \alg accurately recovers the fine-scale quantities of interest. \alg allows for: (i) the output to be sampled at all spatio-temporal resolutions, (ii) a set of Partial Differential Equation (PDE) constraints to be imposed, and (iii) training on fixed-size inputs on arbitrarily sized spatio-temporal domains owing to its fully convolutional encoder.

We empirically study the performance of \alg on the task of super-resolution of turbulent flows in the Rayleigh–Bénard convection problem. Across a diverse set of evaluation metrics, we show that \alg significantly outperforms existing baselines. Furthermore, we provide a large scale implementation of \alg and show that it efficiently scales across large clusters, achieving 96.80\% scaling efficiency on up to 128 GPUs and a training time of less than 4 minutes. We provide an open-source implementation of our method that supports arbitrary combinations of PDE constraints \footnote{\scriptsize{source code} available: \url{https://github.com/maxjiang93/space_time_pde}}.

\end{abstract}
\section{Introduction}
{\color{black}}

In recent years, along with the significant growth of computational resources, there has been an increasing attempt towards accurately resolving the wide range of length and time scales present within different physical systems. These length and time scales often differ by several orders of magnitude. Such time and length scale disparities are commonly observed in different physical phenomena such as turbulent flows, convective diffusive systems, subsurface systems, multiphase flows, chemical processes, and climate systems \citep{Hu2007,Hurrell2009,Bagchi2010,Liu1995,Szymczak2013,Qin2020,Quan2011,RIAZ2006,Esmaeilzadeh2020,Esmaeilzadeh2019}. Yet many key physical quantities of interest are highly dependent on correctly resolving such fine scales, such as the energy spectrum in turbulent flows.

From a numerical perspective, resolving the wide range of spatio-temporal scales within such physical systems is challenging since extremely small spatial and temporal numerical stencils would be required. In order to alleviate the computational burden of fully resolving such a wide range of spatial and temporal scales, multiscale computational approaches have been developed. For instance, in the subsurface flow problem, the main idea of the multiscale approach is to build a set of operators that map between the unknowns associated with the computational cells in a fine-grid and the unknowns on a coarser grid. The operators are computed numerically by solving localized flow problems. The multiscale basis functions have subgrid-scale resolutions, ensuring that the fine-scale heterogeneity is correctly accounted for in a systematic manner \citep{Jenny2006,Jenny2005,Jenny2003}. Similarly, in turbulent flows \citep{Bramkamp2004,Modest2005}, climate forecast \citep{Shen2013}, multiphase systems \citep{Luo2009,Bi2004}, reactive transport \citep{Bauer2001,Odman1991}, complex fluids \citep{Ren2005} and biological systems \citep{Perdikaris2016}, multiscale approaches attempt to relate the coarse-scale solutions to the fine-scale local solutions taking into account the fine-scale physics and dynamics. In these multiscale approaches, although the small scale physics and dynamics are being captured and accounted for at larger scales, the fine-grid solutions are extremely costly and hence impractical to solve for. Reconstructing the fine-scale subgrid solutions by having coarse-scale solutions in space and/or time still remains a challenge.

In this manuscript, we refer to such a process of reconstructing the fine-scale subgrid solutions from the coarse-scale solutions as \textit{super-resolution}, where the high-resolution solutions are reconstructed using the low-resolution physical solutions in space and/or time. One key insight is that there are inherent statistical correlations between such pairs of low-resolution and high-resolution solutions that can be leveraged to reconstruct high-resolution solutions from the low-resolution inputs. Furthermore, this super-resolution process can be effectively modeled using deep learning models that learn such statistical correlations in a self-supervised manner from low-resolution and high-resolution pairs that exist in a training dataset. A successful super-resolution model should be able to efficiently represent the high-resolution outputs, effectively scale to large spatio-temporal domains as in a realistic problem setting, and allow for incorporating physical constraints in the form of PDEs to regularize the outputs to physically valid solutions. Furthermore, in order for a learning-based methodology to be applied to real problems such as Computational Fluid Dynamics (CFD) simulations and climate modeling, the methodology needs to address the High Performance Computing (HPC) challenges of scalability to large scale computing systems, for both the training and inference stages.

To this end, we propose \alg, a novel physics-constrained, deep learning based super-resolution framework to generate continuous (grid-free) spatio-temporal solutions from the low-resolution inputs. \alg first maps the low-resolution inputs to a localized latent context grid using a convolutional encoder, which can then be continuously queried at arbitrary resolutions. In summary, our main contributions are as follows:
\begin{itemize}
\setlength\itemsep{0.1em}
    \item We propose \alg, a novel and efficient physics-constrained deep learning model for super-resolution tasks.
    \item We implement a set of physics-based metrics that allow for an objective assessment of the reconstruction quality of the super-resolved high-resolution turbulent flows.
    \item We empirically assess the effectiveness of the \alg framework on the task of super-resolving turbulent flows in the Rayleigh–Bénard convection problem, showing a consistent and significant improvement in recovering key physical quantities of interest in super-resolved solutions over competing baselines.
    \item We demonstrate the scalability of our framework to larger and more challenging problems by providing a large scale implementation of \alg that scales to up to 128 GPUs while retaining a $\sim97\%$ scaling efficiency.
\end{itemize}

\section{Related Works}

Recently, deep learning models have been applied for studying fluid flow problems in different applications. In particular, a certain class of research works has studied flow problems on a grid representation where dynamic flow properties (\textit{e.g.}, velocity, pressure, etc.) could be considered as \textit{structured} data similar to images. For instance, \citet{guo2016convolutional} used convolutional architectures for approximating non-uniform steady laminar flow around 2D and 3D objects.  Similarly, \citet{bhatnagar2019prediction} used convolutional neural networks for prediction of the flow fields around airfoil objects and for studying aerodynamic forces in a turbulent flow regime. \citet{Zabaras} used Bayesian deep convolutional neural networks to build surrogate flow models for the purpose of uncertainty quantification for flow applications in heterogeneous porous media. \\
\indent
Alternatively, a different area of research has attempted to solve PDEs governing different physical phenomena by using simple fully connected deep neural networks. In order to enforce physical constraints, the PDEs under consideration are added as a term to the loss function at the training stage. For instance, \citet{Weinan} used fully connected deep neural networks with skip connections for solving different classes of PDEs such as Poisson equations and eigenvalue problems. \citet{bar2019unsupervised} used fully connected deep neural networks for solving forward and inverse problems in the context of electrical impedance tomography governed by elliptical PDEs. \citet{long2017pde} proposed a framework to uncover unknown physics in the form of PDEs by learning constrained convolutional kernels.  \citet{raissi2019physics} applied fully connected neural networks for solving PDEs in a forward or inverse setup. \citet{smith2020eikonet} deployed residual neural networks to solve the Eikonal PDE equation and perform ray tracing for an application in earthquake hypo-center inversion and seismic tomography. Using their proposed framework they solved PDEs in different contexts such as fluids, quantum mechanics, seismology, reaction-diffusion systems, and the propagation of nonlinear shallow-water waves.\\
\indent
More recently, alternative representations for spatial functions have been explored. \citet{Anima2020} proposed using message passing on graph networks in order to map between input data of PDEs and their solutions. They showed that the learned networks can generalize between different PDE approximation methods (\textit{e.g.}, Finite Difference and Finite Element methods) and between approximations that correspond to different levels of discretization. Similar to our \alg framework in using a latent context grid that can be continuously decoded into an output field, \citet{jiang2020lig} used such representations for the computer vision task of reconstructing 3D scenes, where latent context grids are decoded into continuous implicit functions that represent the surfaces of different geometries.\\
\indent
In the area of fluid dynamics, for turbulent flow predictions, \citet{wang2019towards} developed a physics-informed deep learning framework. They introduced the use of trainable spectral filters coupled with Reynolds-averaged Navier-Stokes (RANS) and Large Eddy Simulation (LES) models followed by a convolutional architecture in order to predict turbulent flows. \citet{jiang2020enforcing} presented a differentiable physics layer within the neural networks using spectral methods in order to enforce hard linear constraints. By performing a linear projection in the spectral space, they were able to enforce a divergence-free condition for the velocity. Accordingly, they could conserve mass within the physical system both locally and globally, allowing the super-resolution of turbulent flows to be statistically accurate.\\
\indent
In computer vision applications, the \textit{super-resolution} (SR) process has been proposed in the context of reconstructing high-resolution (HR) videos/images from the low-resolution (LR) ones. In the literature, different classical super-resolution methods have been proposed such as prediction-based methods \citep{Irani1991,Keys1981}, edge-based methods \citep{Freedman2011,Sun2008}, statistical methods \citep{KwangInKim2010,ZhiweiXiong2010}, patch-based methods \citep{Freeman2002,HongChang2004} and sparse
representation methods \citep{JianchaoYang2010,JianchaoYang2008}. Most recently, deep learning based super-resolution models have also been actively explored and differ mainly in their types of specific intended application, architectures, loss functions, and learning principles \citep{Lai2017,Ahn2018,Johnson2016,Bulat2018,Lim2017,Wang2018}.

\section{Preliminaries}
In this section, we provide a detailed explanation of the problem set up, notations, data sets, along with the learning paradigm and evaluation metrics.
\subsection{Problem Setup}
Consider a space and time dependent partial differential equation (\PDE) with a unique solution, written in the general form as
\begin{subequations}
\begin{alignat}{3}
\! & \Operator\,\bm{y}={s} \quad \forall \bm{x}\in \Domain\,, & \qquad \label{eq:PDEmain1} \\
\! & \BOperator\bm{y}=b\quad\,\forall \bm{x}\in \partial\Domain\,, & \qquad \label{eq:PDEmain2}
\end{alignat}\label{eq:PDE}
\end{subequations}
defined on a spatio-temporal domain 
$\Domain:=\DomainP\times[0,T]\in\Real^{d+1}$ where $\DomainP$ represents a $d$ dimensional spatial domain, and $[0,T]$ represents the one dimensional temporal domain. Here the $\Operator$ is the operator defining the \PDE within the domain ($\Domain$), $s$ is the source function, $b$ is the boundary condition function, $\BOperator$ is the operator defining the \PDE on the boundary ($\partial\Domain$), and the solution to the \PDE in Eqn.~\ref{eq:PDE} is $\bm{y}:\Omega\rightarrow \Real^m$.

For a partial differential equation (Eqn.~\ref{eq:PDE}), a problem instance is realized with a given source and boundary condition functions $s$ and $b$. For a pair $\left(s,b\right)$, we consider $\GOperator$ to be an operator that generates the high-resolution solution $\bm{y}_\GOperator$, \textit{i.e.}, $\bm{y}_\GOperator=\GOperator(s,b)$, and $\LOperator$ to be an operator that produces the low-resolution solution $\bm{y}_\LOperator$, \textit{i.e.}, $\bm{y}_\LOperator=\LOperator(s,b,\bm{y}_\GOperator)$. 
Consider a compact normed space of operators $\Pi_\F$, a set of operators $\F\in\Pi_\F$, mapping low-resolution solutions to high-resolution ones. 
For a given compact normed space of functions $\Pi_{u,b}$, let $\varepsilon\left(\Pi_\F,\Pi_{s,b}\right)$ denote the approximation gap with respect to $\Pi_\F$, \textit{i.e.},
\begin{equation}\label{eq:maxmin}
\begin{split}
    \varepsilon&\left(\Pi_\F,\Pi_{u,b}\right):=\\
    &\quad\quad\max_{\left(b,u\right)\in\Pi_{u,b}}\min_{\F\in\Pi_\F}\|\GOperator(s,b)-\F\left(\LOperator(s,b,\bm{y}_\GOperator)\right)\|\,,
\end{split}
\end{equation}
\noindent with continuous approximation error $\|\GOperator(s,b)-\F\left(\LOperator(s,b,\bm{y})\right)\|$ in $
F,s,b$. Here the norm is with respect to a desired $L^p(\mu)$ space where $p\geq 1$ and  $\mu$ is a preferred measure\footnote{Note that in the max-min game of the Eqn.~\ref{eq:maxmin}, the action of the environment player, $(s,b)$, is revealed to the approximator player to choose $\F$. Therefore, since the game is in the favor of the approximating player, the value of the game, $\varepsilon\left(\Pi_\F,\Pi_{s,b}\right)$ can be much smaller than its min-max counterpart. We require the min-max value to be desirably small when we aim to have a single model $\F$ to perform well on a set of designated problems.}. In this work, given a problem instance specified with $(s,b)$, we are interested in learning $\F$ using parametric models. In order to map a low-resolution solution to its corresponding super-resolution with low approximation error, we require $\varepsilon\left(\Pi_\F,\Pi_{s,b}\right)$ to be small and comparable with a desired error level. Therefore, given a problem set $\Pi_{u,b}$, a proper and rich class of operators, $\Pi_\F$, allows for a desirable approximation error. 

For a given PDE in Eqn~\ref{eq:PDE}, defined on a hyper-rectangular domain in $\Real^{d+1}$, we consider a spatio-temporal grid of resolutions $(n_1,n_2,\ldots, n_{d+1})$ with $n_{d+1}$ denoting the number of discretized points in time domain, and a data set $\D:=\{(\bm{x}^i, \bm{y}^i)\}_{i=1}^{n_1\cdot n_2\cdot\ldots\cdot n_{d+1}}$, of points and their solution values.

\begin{figure}[h!]
    \centering
    \psfrag{t1}{$n_1$}
        \includegraphics[width=0.8\linewidth]{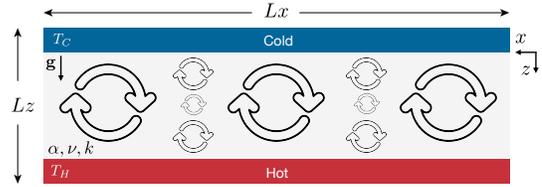}
        \psfrag{n1}{$n_1$}
        \caption{\small Configuration of the Rayleigh–Bénard instability problem - Cold and hot plates have temperatures of $T_C$ and $T_H$ and separated with a distance of $L$. Flow parameters $\alpha$, $\nu$, $\kappa$ respectively are the thermal expansion coefficient, kinematic viscosity, and thermal diffusivity. $g$ is the gravity acceleration in the $z-$direction, and $L_x$, $L_z$ respectively are the length of the plates and their separation distance. $x$ and $z$ respectively refer to the coordinates of the Cartesian frame of reference.}
    \label{fig:reayleighBenard}
    \vspace{-1em}
\end{figure}
\vspace{-1em}
\begin{figure}[h!]
    \centering
    \psfrag{t1}{$n_1$}
        \includegraphics[width=0.6\linewidth]{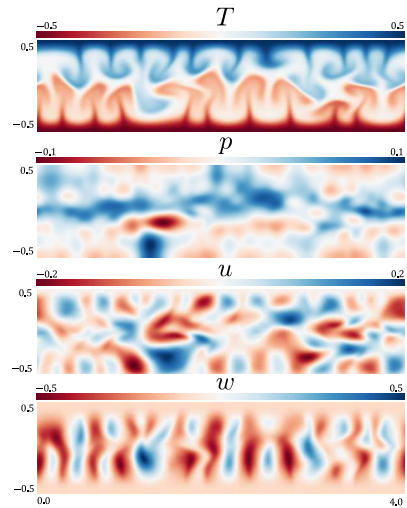}
        \psfrag{n1}{$n_1$}
        \caption{\small An illustration of a typical solution to the Rayleigh–Bénard problem (Eqns.\,\eqref{eq:Rayleigha}-\eqref{eq:Rayleighc}). The contours  respectively show the solution for temperature ($T$), pressure ($p$), and the two velocity components ($u$ and $w$). For this simulation case $Pr = 1$, $Ra = 10^6$, and $L_x = 4L_z = 4$ [$m$]. The spatial resolution is $n_x$ = $4\times n_z$ = $512$, and upon an adaptive time stepping scheme the presented solution above is obtained at $t = 25$ [$s$].}
    \label{fig:reayleighBenard2}
\end{figure}

\subsection{Dataset Overview}\label{ssec:data}
In this work, we generate the dataset as the solution to a classical fluid dynamics system with a chaotic nature. We consider the well-known Rayleigh–Bénard instability problem in 2D where a static bulk fluid (kinematic viscosity $\nu$, thermal diffusivity $\alpha$) is initially occupying the space between two horizontal plates (see Fig. \ref{fig:reayleighBenard}). The lower and upper plates are considered to be respectively hot and cold with temperatures of $T_H$ and $T_C$. Gradually the temperature of the bulk fluid adjacent to the hot (cold) plate increases (decreases) and due to the buoyancy effects and density gradient the bulk fluid ascends (descends), leading to the formation of vortices and growth of flow instability in a chaotic and turbulent regime. The governing partial differential equations for the Rayleigh–Bénard instability problem are
\begin{subequations}
\begin{alignat}{3}
\! & \nabla\cdot\mathbf{u} = 0\,, & \qquad \label{eq:Rayleigha} \\
\! & \frac{\partial \mathbf{u}}{\partial t} + \mathbf{u} \cdot \nabla \mathbf{u} + \nabla p - T\hat{z} - R^* \nabla^2 \mathbf{u} = 0\,, & \qquad & \\
\! & \frac{\partial T}{\partial t} + \mathbf{u} \cdot \nabla T - P^* \nabla^2 T = 0\,, & \qquad \label{eq:Rayleighc}
\end{alignat}
\end{subequations}
where $P^* = (Ra\,Pr)^{-1/2}$ and $R^* = (Ra/Pr)^{-1/2}$ with $Ra$ and $Pr$ being the Rayleigh and Prandtl numbers respectively defined as $Ra = g\alpha\Delta T L^3\nu^{-1} \kappa^{-1}$ and $Pr=\nu\kappa^{-1}$, with $g$, $\alpha$, $\nu$, $\kappa$, $\Delta T$, and $L$ respectively being the gravity acceleration, thermal expansion coefficient, kinematic viscosity, thermal diffusivity, temperature difference between hot and cold plates, and the separation length between the plates. From a physical perspective $Ra$ quantifies the balance between the gravitational forces and viscous damping, and $Pr$ quantifies the ratio between momentum diffusivity and thermal diffusivity (balance between heat convection and conduction).

We use the Dedalus framework \citep{Burns2019} in order to numerically solve the system of Equations \eqref{eq:Rayleigha}-\eqref{eq:Rayleighc} using the spectral method approach. We solve Equations \eqref{eq:Rayleigha}-\eqref{eq:Rayleighc} for a duration of $t_f$ in time with a time step size of $\Delta t$. In a coordinate system of $x$-axis, and $z$-axis, we consider a plate length $L_x$ and a separation distance $L_z$, and discretize the domain with $n_x$ and $n_z $ points respectively in the $x$ and $z$ directions. 

For the simulation cases, we consider Rayleigh and Prandtl numbers respectively in the range of $Ra \in [10^4,\,10^8]$ and $Pr \in [0.1,\,10]$. Upon solving the system of Eqns.\,\eqref{eq:Rayleigha}-\eqref{eq:Rayleighc}, we create a high-resolution Rayleigh–Bénard simulation dataset $\D_H$, unless otherwise mentioned, with a spatial resolution of $n_x$ = $4\times n_z$ = $512$, and a temporal resolution of $n_t$ = $400$ (upon adaptive time stepping). We consider a normalized domain size of unit length in $z$-direction with a domain aspect ratio of 4, \textit{i.e.}, $L_x = 4\times L_z=4$ [$m$], and solve the Rayleigh–Bénard problem for a duration of 50 [$s$]. Then we create a low-resolution dataset $\D_L$, by downsampling the high-resolution data in both space and time. We use downsampling factors of $d_t = 4$ and $d_s = 8$ for creating the low-resolution data in the temporal and spatial dimensions respectively.

\subsection{Evaluation Metrics}\label{ssec:evalMetric}
In this work, we use multiple physical metrics, each accounting for different aspects of the flow field, in order to report the evaluation of the \alg model for super-resolving low-resolution data. As the specific metrics of evaluation, we report the Normalized Mean Absolute Error (NMAE) and R2 score for the physical metrics between the ground truth and the predicted high-resolution data. Such physical metrics are listed in the following.
\begin{itemize}
\setlength\itemsep{0.1em}
\item \textit{Total Kinetic Energy ($E_{tot}$)}\,: the kinetic energy per unit mass associated with the flow is defined as the total kinetic energy and can be expressed as $E_{tot}=\frac{1}{2}\langle u_i u_i\rangle$\,.
\item \textit{Root Mean Square Velocity ($u_{rms}$)}\,: the square root of the scaled total kinetic energy is defined as the root mean square (RMS) velocity as $u_{rms} = \sqrt{({2}/{3})E_{tot}}$\,.
\item \textit{Dissipation ($\varepsilon$)}\,: is the rate at which turbulence kinetic energy is converted into thermal internal energy by molecular viscosity and defined as $\varepsilon = 2\nu\langle S_{ij}S_{ij}\rangle$ with $S_{ij}$ and $\nu$ respectively being the rate of strain tensor and the kinematic viscosity.
\item \textit{Taylor Microscale ($\lambda$)}\,: is the intermediate length scale at which viscous forces significantly affect the dynamics of turbulent eddies and defined as $\lambda = \sqrt{15\,\nu\,u_{rms}^2\,\varepsilon^{-1}}$. Length scales larger than the Taylor microscale (\textit{i.e.}, inertial range) are not strongly affected by viscosity. Below the Taylor microscale (\textit{i.e.}, dissipation range) the turbulent motions are subject to strong viscous forces and kinetic energy is dissipated into heat.
\item \textit{Taylor-scale Reynolds ($Re_\lambda$)}\,: is defined as the ratio of RMS inertial forces to viscous forces and is expressed as $Re_\lambda = u_{rms}\,\lambda\,\nu^{-1}$\,.
\item \textit{Kolmogorov Time ($\tau_\eta$) and Length ($\eta$) Scales}\,: Kolmogorov microscales are the smallest scales in turbulent flows where viscosity dominates and the turbulent kinetic energy dissipates into heat. Kolmogorov time and length scale can be respectively expressed as $\tau_\eta = \sqrt{\nu/\varepsilon}$ and $\eta = \nu^{3/4}\,\varepsilon^{-1/4}$\,.
\item \textit{Turbulent Integral Scale ($L$)}\,: is a measure of the average spatial extent or coherence of the fluctuations and can be expressed as $L = \frac{\pi}{2u^2_{rms}}\int{\frac{E(k)}{k}}dk$\,.
\item \textit{Large Eddy Turnover Time ($T_L$)}\,: is defined as the typical time scale for an eddy of length scale $L$ to undergo significant distortion and is also the typical time scale for the transfer of energy from scale $L$ to smaller scales, since this distortion is the mechanism for energy transfer and expressed as $T_L = L/u_{rms}$\,.
\end{itemize}
\begin{figure*}[t!]
\begin{floatrow}
\ffigbox[.6\textwidth]{%
    \centering
    \includegraphics[width=\linewidth, trim={0 0 0 0}, clip]{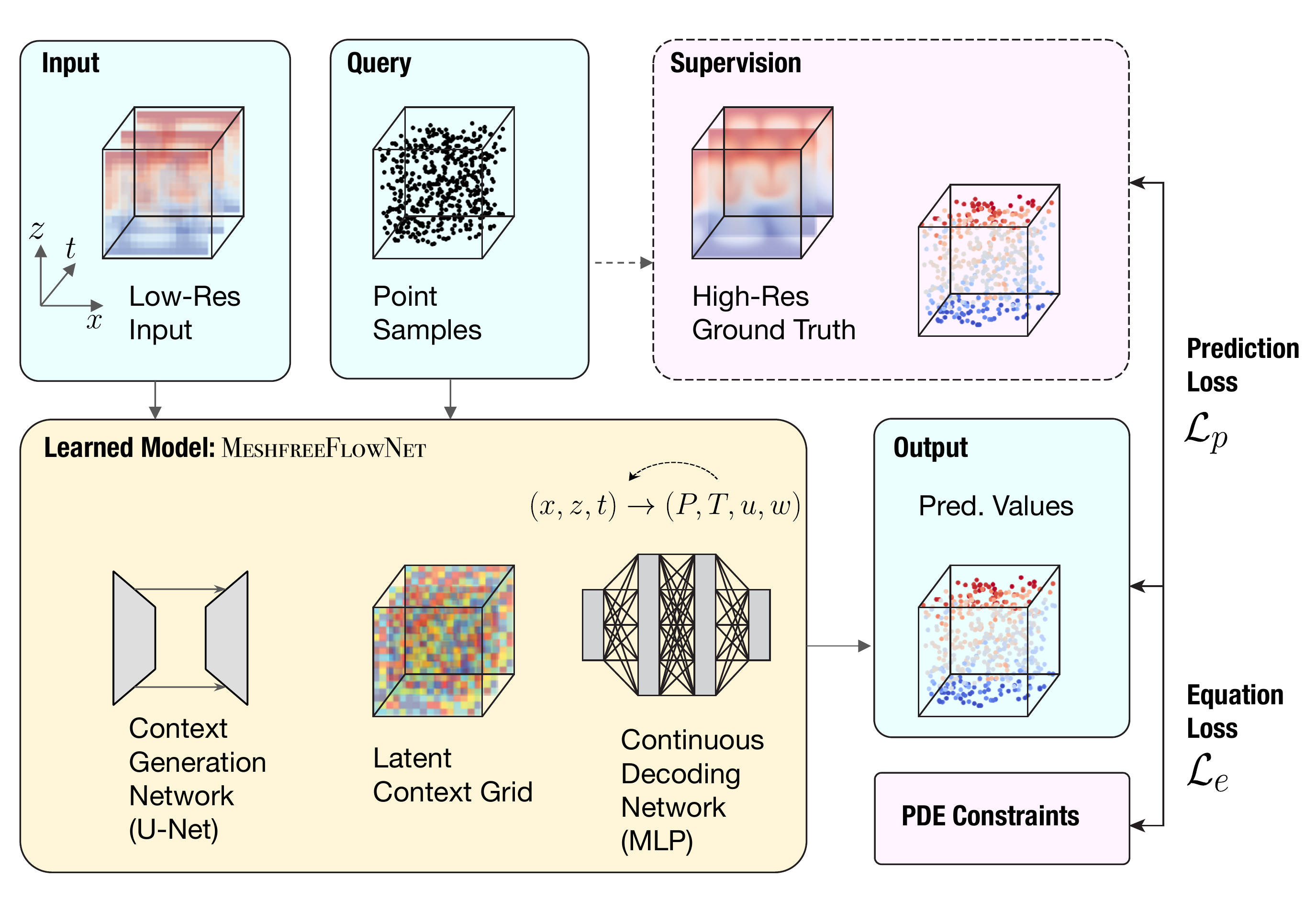}
}{%
  \caption{\small Schematic for the training pipeline for \alg model for continuous space-time super-resolution. An input low-resolution grid is fed to the Context Generation Network that creates a Latent Context Grid. A random set of points in the corresponding space-time domain is sampled to query the Latent Context Grid, and the physical output values at these query locations can be continuously decoded using a Continuous Decoding Network, implemented as a Multilayer Perception. Due to the differentiable nature of the MLP, any partial derivatives of the output physical quantities with respect to the input space-time coordinates can be effectively computed via backpropagation, that can be combined with the PDE constraints to produce an Equation Loss. On the other hand, the predicted value can be contrasted with the ground truth value at these locations produced by interpolating the high-resolution ground truth to produce a Prediction Loss. Gradients from the combined losses can be backpropagated to the network for training.}
  \label{fig:pcsr}
}
\ffigbox[.35\textwidth]{%
    \centering
    \includegraphics[width=\linewidth]{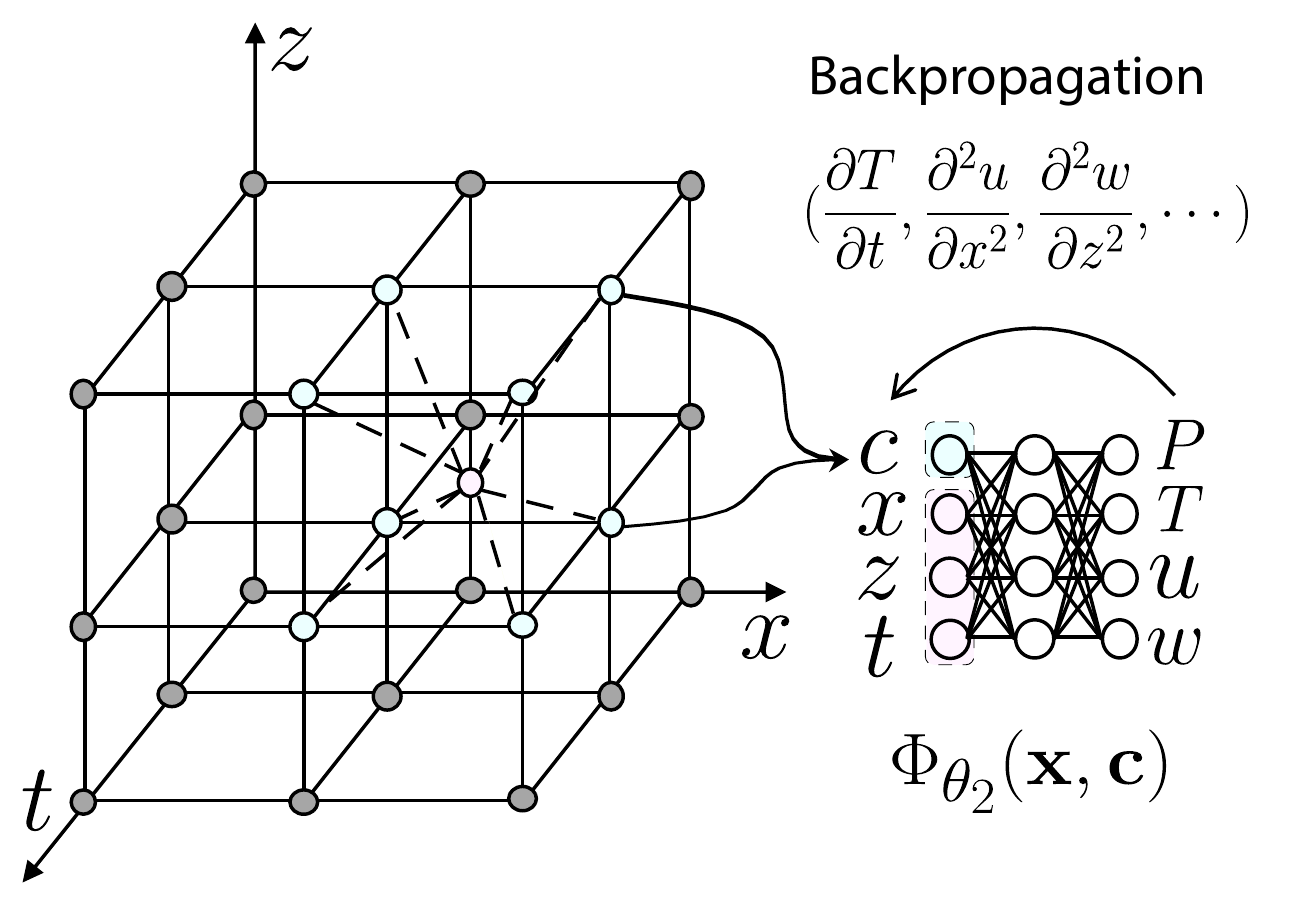}
}{%
  \caption{\small Schematic for the continuous decoding module for \alg. The continuous decoding network is a Multilayer Perception that inputs the spatio-temporal coordinates of a query point, along with a latent context vector, and is decoded into the required physical channels of interest. Since each query point falls into a cell bounded by 8 neighboring vertices, the query is performed 8 times, each using a different latent context vector and a different relative spatio-temporal coordinate with respect to each vertex. The values are then interpolated using trilinear interpolation to get the final value at the query point.}
  \label{fig:cont_dec}
}
\end{floatrow}
\end{figure*}

\subsection{Systems, Platforms and Configuration}\label{ssec:SPC}
In order to further illustrate the feasibility of utilizing our proposed \alg framework for large physical systems requiring processing orders of magnitude more computation, here we study the scalability of \alg. We scale our model on a GPU stack on the Cori supercomputer at the National Energy Research Scientific Computing Center (NERSC). We use data distributed parallelism with synchronous gradient descent, and test performance up to 16 nodes with a total of $128$ GPUs.

In a data-parallel distribution strategy, the model is replicated over all GPUs. At every step, each GPU receives its own random batch of the dataset to calculate the local gradients. Gradients are averaged across all devices with an {\tt all\_reduce} operation. To achieve better scaling efficiency, the communication of one layer's gradients is overlapped with the {\tt backprop} computation of the previous layer. PyTorch {\tt torch.distributed} \citep{torch_distributed} package provides communication primitives for multiprocess parallelism with different collective communication backends. We use NVIDIA's Collective Communications Library (NCCL)~\citep{nccl} backend which gave us the best performance when running on GPUs, both within and across multiple nodes. The PyTorch {\tt torch.nn.parallel.DistributedDataParallel} \citep{torch_ddp} wrapper, which we use for the results in this paper, builds on top of the {\tt torhch.distributed} package to provide efficient data-parallel distributed training. With this setup, we achieve more than $96\%$ throughput efficiency on 16 Cori GPU nodes, 128 GPUs in total. Cori has 8 V100 (Volta) GPUs per node, the GPUs are interconnected with NVLinks in hybrid cube-mesh topology~\citep{cori_gpu_topog}. The nodes are equipped with Mellanox MT27800 (ConnectX-5) EDR InfiniBand network cards.

\section{\alg}
\begin{figure*}[t]
\includegraphics[width=\textwidth]{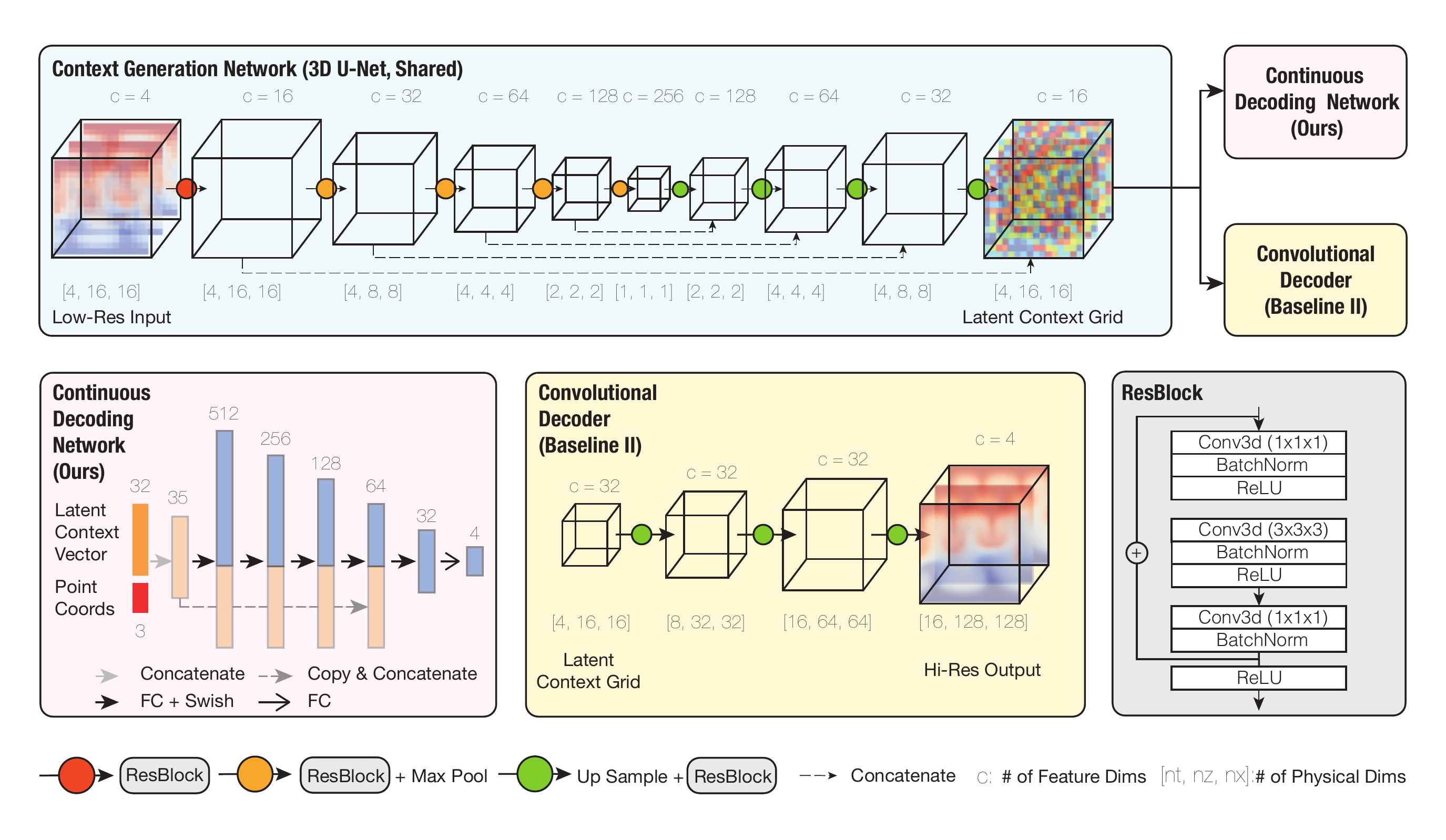}
\caption{\small A schematic for the architecture of the \alg framework. \alg consists of two end-to-end trainable modules: the Context Generation Network, and the Continuous Decoding Network. The Baseline (II) method that we exhaustively compare with share the same 3D U-Net structure in the encoding arm, but instead uses a convolutional decoder for producing the final output. In comparison, the \alg allows for continuous output that can be sampled at arbitrary space-time locations.}
\label{fig:arch}
\end{figure*}
In this work, we propose the \alg framework as a novel computational algorithm for constructing the super-resolution solutions to partial differential equations using their low-resolution counterpart solutions.

A schematic for the training pipeline for the \alg model is presented in Fig.~\ref{fig:pcsr}. The model inputs a low-resolution spatio-temporal grid that can be acquired from a PDE solver, along with query locations that can be any continuous value within the range of the input domain. \alg produces a predicted output vector at each query location. 
\alg consists of two subnetworks, namely the \textit{Context Generation Network}, and the \textit{Continuous Decoding Network}. The Context Generation Network produces a \textit{Latent Context Grid}, which is a learned localized representation of the function. The Latent Context Grid contains latent context vectors $\bm{c}\in\mathbb{R}^{n_c}$ at each grid vertex, which along with spatio-temporal coordinates $\bm{x}:=\{x, z, t\}$, can be concatenated and fed into the Continuous Decoding Network, modeled as a Multilayer Preceptron (MLP), to generate the physical outputs $\bm{y}\in\mathbb{R}^m$. For training the model, two loss terms, namely the prediction loss and the equation loss are used. We present a more detailed architectural breakdown for our method and our baseline method in Fig.~\ref{fig:arch}. We discuss the details of the Context Generation Network (Sec.~\ref{ssec:cgn}), Continuous Decoding Network (Sec.~\ref{ssec:cdn}), and loss functions (Sec.~\ref{ssec:loss}) in the following subsections.

\renewcommand{\arraystretch}{1}
\begin{table*}[h!]
\centering
\resizebox{0.8\columnwidth}{!}{%
\begin{tabular}{c|ccccccccc|c}
                                    & \multicolumn{9}{c|}{\makecell{100$\times$NMAE\\(R2)}}                                              &                       \\ 
$\gamma$ & $E_{tot}$ & $u_{rms}$ & $\varepsilon$ & $\lambda$ & $Re_{\lambda}$ & $\tau_{\eta}$ & $\eta$ & $L$ & $T_L$ & avg. R2 \\ \hline
0      &\makecell{0.667\\(0.9991)}    &\makecell{0.768\\(0.9987)}   &\makecell{0.666\\(0.9991)}   &\makecell{0.545\\(0.9985)}   &\makecell{0.444\\(0.9989)}   &\makecell{0.753\\(0.9981)}   &\makecell{0.752\\(0.9984)}   &\makecell{0.837\\(0.9968)}   &\makecell{0.455\\(0.995)} &  0.9986   \\\cline{1-1} 
0.0125 &\makecell{0.650\\(0.9990)}    &\makecell{0.616\\(0.9992)}   &\makecell{0.639\\(0.9991)}   &\makecell{0.457\\(0.9989)}   &\makecell{0.435\\(0.9986)}   &\makecell{0.589\\(0.9988)}  &\makecell{0.588\\(0.9990)}   &\makecell{\bb{0.670}\\(0.9982)}   &\makecell{0.432\\(0.9994)}   &0.9989                       \\ \cline{1-1}
0.025  &\makecell{0.671\\(0.9993)}&\makecell{0.699\\(0.9992)}&\makecell{0.671\\(0.9993)}&\makecell{0.454\\(0.9990)}&\makecell{\bb{0.332}\\(0.9994)}&\makecell{0.705\\(0.9985)}&\makecell{0.698\\(0.9988)}&\makecell{0.781\\(0.9973)}&\makecell{0.430\\(0.9996)}&0.9989                     \\ \cline{1-1}
\bb{$\gamma^*$\,$=$\,0.05}   &\makecell{\bb{0.621}\\(0.9993)}&\makecell{\bb{0.603}\\(0.9992)}&\makecell{\bb{0.617}\\(0.9993)}&\makecell{\bb{0.431}\\(0.9992)}&\makecell{0.429\\(0.9989)}&\makecell{\bb{0.461}\\(0.9994)}&\makecell{\bb{0.483}\\(0.9994)}&\makecell{0.857\\(0.9972)}&\makecell{\bb{0.418}\\(0.9996)}&\bb{0.9991}    \\  \cline{1-1}
0.1    &\makecell{3.209\\(0.9894)}&\makecell{1.790\\(0.9954)}&\makecell{3.015\\(0.9907)}&\makecell{1.024\\(0.9907)}&\makecell{1.013\\(0.9917)}&\makecell{1.628\\(0.9902)}&\makecell{1.636\\(0.9925)}&\makecell{1.970\\(0.9900)}&\makecell{2.443\\(0.9922)}&0.9914
\\  \cline{1-1}
0.2    &\makecell{1.396\\(0.9979)}&\makecell{5.059\\(0.9679)}&\makecell{1.373\\(0.9978)}&\makecell{3.964\\(0.8835)}&\makecell{3.553\\(0.8781)}&\makecell{4.329\\(0.9270)}&\makecell{4.415\\(0.9436)}&\makecell{3.589\\(0.9739)}&\makecell{1.636\\(0.9962)}&0.9518                       \\ \cline{1-1}
0.4    &\makecell{3.560\\(0.9854)}&\makecell{13.270\\(0.7909)}&\makecell{3.533\\(0.9854)}&\makecell{10.119\\(-0.177)}&\makecell{10.056\\(-0.366)}&\makecell{9.9149\\(0.466)}&\makecell{10.392\\(0.5923)}&\makecell{5.787\\(0.9343)}&\makecell{1.751\\(0.9917)}&0.5780                       \\ \cline{1-1}
0.8    &\makecell{11.433\\(0.8507)}&\makecell{15.727\\(0.6805)}&\makecell{11.171\\(0.8701)}&\makecell{13.502\\(-1.610)}&\makecell{14.664\\(-2.556)}&\makecell{11.778\\(0.0830)}&\makecell{12.385\\(0.3005)}&\makecell{2.092\\(0.9819)}&\makecell{8.033\\(0.9231)}&0.0582                       \\\cline{1-1}
1.0    &\makecell{13.617\\(0.7954)}&\makecell{17.970\\(0.6031)}&\makecell{13.441\\(0.8206)}&\makecell{15.080\\(-2.536)}&\makecell{16.508\\(-3.746)}&\makecell{12.868\\(-0.121)}&\makecell{13.631\\(0.1487)}&\makecell{6.165\\(0.9179)}&\makecell{8.546\\(0.9154)}&-0.2447  

\end{tabular}
}
\vspace{1 pt}
\caption{\small Normalized Mean Absolute Error (NMAE) and R2-score of the flow-based evaluation metrics evaluated for the predicted vs. the ground truth high-resolution validation data. $\gamma$ refers to the coefficient of the Equation loss ($\mathcal{L}_e$) in the total loss function (Eqn.\,\ref{eq:loss}).}
\label{table:gamma}
\end{table*}

\subsection{Context Generation Network}\label{ssec:cgn}
The \textit{Context Generation Network} is a convolutional encoder that produces a Latent Context Grid from the low-resolution physical input $\D_L$. Denote this network as
\vspace{-.1em}
\begin{equation}
    \G = \Psi_{\theta_{1}}(\D_L)\,,
\end{equation}
where $\theta_1$ is the set of learnable parameters associated with this network, $\G$ is the generated context grid, $\G\in \mathbb{R}^{n_1\times n_2 \times \cdots \times n_{d+1} \times n_c}$ where $n_c$ is the number of latent channels or the dimensions of each latent context vector. It is worth noting that the function $\Psi_{\theta_1}$ can be applied on a small fixed-size sub-window of $\Domain$, since the model is fully-convolutional. By applying it to the input in a fully-convolutional manner, the size of the domain at test time (both spatial and temporal sizes) can be orders of magnitude greater than the dimensions of $\D_L$. 

In this work, we implement the Context Generation Network as a 3D variant of the \textit{U-Net} architecture which was originally proposed by~\citet{Ronneberger2015}. U-Net has successfully been applied for different computer vision tasks that involve dense localized predictions such as image style transfer \citep{Zhang2018},
image segmentation \citep{Lian2019,Chen2019}, image enhancement \citep{Chen2018}, image coloring 
\citep{Liu2019,Fang2019}, and image generation \citep{Esser2018,Zhang}

Different from the original U-Net architecture, we replace the 2D convolutions with 3D counterparts and utilize residue blocks instead of individual convolution layers for better training properties of the network. The U-Net comprises of a contractive part followed by an expansive part. The contractive part is composed of multiple stages of convolutional residue blocks, each followed by a max-pooling layer (of stride 2). Each residue block consists of 3 convolution layers (1x1, 3x3, 1x1) interleaved with batch normalization layers and ReLU activation. The expansive part mirrors the contractive part, replacing max-pooling with nearest neighbor upsampling. In between the layers with similar grid sizes within the contractive and expansive parts, a skip connection concatenates the features from the contractive layer with the features in the expansive layer as the input to the subsequent layers in the contractive part in order to preserve the localized contextual information.

\renewcommand{\arraystretch}{1}
\begin{table*}[h!]
\centering
\resizebox{0.8\columnwidth}{!}{
\begin{tabular}{c|ccccccccc|c}
& \multicolumn{9}{c|}{\makecell{100$\times$NMAE\\(R2)}}                    & \\ 
Model& $E_{tot}$ & $u_{rms}$ & $\varepsilon$ & $\lambda$ & $Re_{\lambda}$ & $\tau_{\eta}$ & $\eta$ & $L$ & $T_L$ & avg. R2 \\ \hline
Baseline (I)  &\makecell{69.6360\\(-19.7894)}    &\makecell{3470.132\\(-57.7717)}   &\makecell{76.338\\(-14)}   &\makecell{78.164\\(-11096)}   &\makecell{75.729\\(-4934)}   &\makecell{77.410\\(-12599)}  &\makecell{122.55\\(-3147)}    &\makecell{137.376\\(-0.5190)}   &\makecell{109.398\\(-3.0092)}   &{-3541} \\ \cline{1-1}
Baseline (II)  & \makecell{6.489\\(0.9557)}    &\makecell{8.769\\(0.8967)}   &\makecell{6.144\\(0.9593)}   &\makecell{3.903\\(0.9490)}   &\makecell{2.489\\(0.9711)}   &\makecell{5.584\\(0.9382)}  &\makecell{6.019\\(0.94019)}    &\makecell{2.902\\(0.9597)}   &\makecell{5.076\\(0.9644)}   &{0.9482} \\ \cline{1-1}
\alg,\,$\gamma=0$ &\makecell{0.667\\(0.9991)}    &\makecell{0.768\\(0.9987)}   &\makecell{0.666\\(0.9991)}   &\makecell{0.545\\(0.9985)}   &\makecell{0.444\\(0.9989)}   &\makecell{0.753\\(0.9981)}   &\makecell{0.752\\(0.9984)}   &\makecell{\bb{0.837}\\(0.9968)}   &\makecell{0.455\\(0.995)} &  0.9986 \\ \cline{1-1}
\alg,\,$\gamma=\gamma^*$  &\makecell{\bb{0.621}\\(0.9993)}&\makecell{\bb{0.603}\\(0.9992)}&\makecell{\bb{0.617}\\(0.9993)}&\makecell{\bb{0.431}\\(0.9992)}&\makecell{\bb{0.429}\\(0.9989)}&\makecell{\bb{0.461}\\(0.9994)}&\makecell{\bb{0.483}\\(0.9994)}&\makecell{0.857\\(0.9972)}&\makecell{\bb{0.418}\\(0.9996)}&\bb{0.9991}     
\end{tabular}
}
\vspace{1 pt}
\caption{\small Comparison between the performance of \alg framework for super-resolving the low-resolution data vs. two baseline models.}
\label{tab:baselines}
\end{table*}

\renewcommand{\arraystretch}{1}
\begin{table*}[h!]
\centering
\resizebox{0.8\columnwidth}{!}{
\begin{tabular}{c|ccccccccc|c}
& \multicolumn{9}{c|}{\makecell{100$\times$NMAE\\(R2)}}                                       & \multicolumn{1}{c}{} \\
\makecell{$\#$ Dataset(s)\\($\gamma=\gamma^*$)} & $E_{tot}$ & $u_{rms}$ & $\varepsilon$ & $\lambda$ & $Re_{\lambda}$ & $\tau_{\eta}$ & $\eta$ & $L$ & $T_L$ & avg. R2 \\ \hline
1 &\makecell{0.621\\(0.9993)}&\makecell{0.603\\(0.9992)}&\makecell{0.617\\(0.9993)}&\makecell{0.431\\(0.9992)}&\makecell{0.429\\(0.9989)}&\makecell{\bb{0.461}\\(0.9994)}&\makecell{\bb{0.483}\\(0.9994)}&\makecell{0.857\\(0.9972)}&\makecell{0.418\\(0.9996)}&0.9991                      \\ \cline{1-1}
10 &
       \makecell{\bb{0.609}\\(0.9995)}&\makecell{\bb{0.599}\\(0.9993)}&
       \makecell{\bb{0.603}\\(0.9991)}&\makecell{\bb{0.428}\\(0.9991)}&
       \makecell{\bb{0.411}\\(0.9987)}&\makecell{0.470\\(0.9990)}&
       \makecell{0.497\\(0.9991)}&\makecell{\bb{0.673}\\(0.9998)}&
       \makecell{\bb{0.345}\\(0.9998)}& \bb{0.9993}                     
\end{tabular}
}
\vspace{1 pt}
\caption{\small For a \alg model that has been respectively trained on 1 dataset and 10 datasets each having a different initial condition, the super-resolution performance evaluation is reported on a dataset with an unseen initial condition.}
\label{table:initialCond}
\end{table*}

\subsection{Continuous Decoding Network}\label{ssec:cdn}

One unique property of our super-resolution methodology is that the output is continuous instead of discrete. This removes the limitations in output resolution, and additionally, it allows for an effective computation of the gradients of predicted output physical quantities, enabling an easy way of enforcing PDE-based physical constraints.

The continuous decoding network can be implemented using a simple Multilayer Perceptron, see Fig. \ref{fig:cont_dec}. For each query, denote the spatio-temporal query location to be $\bm{x}_i$ and the latent context grid to be $\G:= \{(\bm{x}_j, \bm{c}_j) ; j \leq ||\G||\}$, where $(\bm{x}_j, \bm{c}_j)$ are the spatio-temporal coordinates and the latent context vector for the $j$-th vertex of the grid. Denote $\mathcal{N}_i$ as the set of neighboring vertices that bound $x_i$, where for a ($d$+1) dimensional spatio-temporal grid $||\mathcal{N}_i||=2^{d+1}$. Denote the continuous decoding network, implemented as a Multilayer Perception as
\begin{equation}
\Phi_{\theta_{2}}(\bm{x}, \bm{c})\,,
\end{equation}
where $\theta_{2}$ is the set of trainable parameters of the Multilayer Perception 
network. The query value at $\bm{x}$ with respect to the shared network $\Phi_{\theta_{2}}(\bm{x})$ and the latent context grid $\G$ can be calculated as
\begin{align}
    \mathcal{C}(\bm{x}_i, \G, \Phi_{\theta_{2}}) = \sum_{j\in \mathcal{N}_i} w_j \Phi_{\theta_{2}}(\frac{\bm{x}_i-\bm{x}_j}{\Delta \bm{x}}, \bm{c}_j)\,,
\end{align}
where $\sum_{j\in \mathcal{N}_i} w_j=1$, $w_j$ is the trilinear interpolation weight with respect to the bounding vertex $j$, $\Delta \bm{x}:=\{\Delta x, \Delta z, \Delta t\}$ is the stencil size corresponding to the discretization grid vertices. 

Since the Continuous Decoding Network is implemented as an MLP, arbitrary spatio-temporal derivatives of the output quantities: $\Gamma\bm{y}$ can be effectively computed via backpropagation through the MLP. Denote the approximation of the derivative operator to be $\Gamma_{\Phi}$. We combine the partial derivatives to compute the equation loss as the norm of the residue of the governing equations. In the continuous decoding network we consider two infinitely differentiable activation functions namely Softplus and Swish, where we have found the results obtained by Swish to outperform the ones obtained by Softplus. 

\subsection{Loss Function}\label{ssec:loss}
We use a weighted combination of two losses to train our \alg network: the norm of the difference between the predicted physical outputs and the ground truth physical outputs, which we refer to as \textit{Prediction Loss}, and the norm of the residues of the governing PDEs, which we refer to as \textit{Equation Loss}. 

Denote the set of sample locations within a mini-batch $\mathcal{B}$ of training samples to be $\{(\bm{x}_j^i, \bm{y}_j^i, \D_L^i); i \in \mathcal{B}, j \in \mathcal{B}^i\}$ where $\mathcal{B}^i$ is the mini-batch of point samples for the $i$-th low-resolution input, $\bm{y}$ is the vector that represents the ground truth physical output quantities. In the case of the Rayleigh-Bénard Convection example in this study, we have $\bm{y}:=\{P, T, u, v\}$ where $P, T, u, v$ are the pressure, temperature, x-velocity and y-velocity terms respectively. The super-resolution for the learned model queried at $\bm{x}$ conditioning on low-resolution input $\D_L$ is
\begin{align}
    \hat{\bm{y}}(\bm{x}, \D_L; \theta_1, \theta_2) &= \C(x, \Psi_{\theta_1}(\D_L), \Phi_{\theta_2})\,,
\end{align}
where $\hat{\bm{y}}$ is the predicted output vector. The \textit{prediction loss} $\mathcal{L}_p$ for a mini-batch can be formulated as
\begin{align}
    \mathcal{L}_p = \frac{1}{|\mathcal{B}|}\sum_{i\in\mathcal{B}}\frac{1}{||B^i||}\sum_{j\in\mathcal{B}^i}||\bm{y}_{j}^{i} - \hat{\bm{y}}_j^i||_l\,,
\end{align}
where $||\cdot||_l$ is the Frobenius $l$-norm of the difference. We use the L1 Norm for computing the prediction loss. The \textit{Equation loss} $\mathcal{L}_e$ for a mini-batch can be formulated as
\begin{align}
    \mathcal{L}_e = \frac{1}{|\mathcal{B}|}\sum_{i\in\mathcal{B}}\frac{1}{||B^i||}\sum_{j\in\mathcal{B}^i}||\Gamma_{\Phi}\hat{\bm{y}}_j^i-s||_l\,,
\end{align}
which is the norm of the PDE equation residue, using the PDE definition in Eqn.~\ref{eq:PDE}. Finally, a single loss term for training the network can be represented as a weighted sum of the two loss terms as
\begin{equation}
    \mathcal{L}=\mathcal{L}_p + \gamma \mathcal{L}_e\,,
    \label{eq:loss}
\end{equation}
where $\gamma$ is a hyperparameter for weighting the equation loss.
\renewcommand{\arraystretch}{1}
\begin{table*}[h!]
\centering
\resizebox{0.8\columnwidth}{!}{
\begin{tabular}{c|ccccccccc|c}
& \multicolumn{9}{c|}{\makecell{100$\times$NMAE\\(R2)}}     
& \multicolumn{1}{c}{} \\
$Ra$ & $E_{tot}$ & $u_{rms}$ & $\varepsilon$ & $\lambda$ & $Re_{\lambda}$ & $\tau_{\eta}$ & $\eta$ & $L$ & $T_L$ & avg. R2 \\ \hline

$1\times10^4$ 
&\makecell{1.180\\(0.9973)} & \makecell{0.401\\(0.99963)} &\makecell{0.709\\(0.9992)}& \makecell{44.6435\\(0.2038)} & \makecell{1.357\\(0.9977)} & \makecell{2.983\\(0.4588)} &\makecell{2.374\\(0.8726)} 
&\makecell{70.34\\(0.0104)}
&\makecell{0.137\\(0.9996)}
&  0.7266                    \\ \cline{1-1}

$1\times10^5$ & \makecell{0.670\\(0.9987)} &\makecell{0.537\\(0.9995)}     &\makecell{0.611\\(0.9990)} & \makecell{0.450\\(0.9992)} &\makecell{0.472\\(0.9989)}& \makecell{0.589\\(0.9992)} & \makecell{0.574\\(0.9993)} & \makecell{0.713\\(0.9990)} &\makecell{0.331\\(0.9996)}    &  0.9992                    \\ \cline{1-1}

$5\times10^6$ & \makecell{1.043\\(0.9981)} &\makecell{1.0601\\(0.9927)}  &\makecell{1.017\\(0.9963)} & \makecell{1.054\\(0.9963)} &\makecell{0.954\\(0.9961)}& \makecell{1.441\\(0.9921)} & \makecell{1.692\\(0.9924)} & \makecell{0.792\\(0.9934)} &\makecell{0.441\\(0.993)}    &  0.9997                    \\ \cline{1-1}

$1\times10^7$     & \makecell{1.395\\(0.9958)} &\makecell{3.044\\(0.9743)}     &\makecell{1.405\\(0.9957)} & \makecell{2.767\\(0.9691)} &\makecell{1.821\\(0.9862)}& \makecell{2.900\\(0.9714)} & \makecell{2.924\\(0.9725)} & \makecell{2.9817\\(0.9749)} &\makecell{1.077\\(0.9969)}    & 0.9819                      \\ \cline{1-1}

$1\times10^8$     & \makecell{3.184\\(0.9870)} &\makecell{4.454\\(0.9664)}     &\makecell{3.221\\(0.9869)} & \makecell{6.362\\(0.8826)} &\makecell{7.471\\(0.8110)}& \makecell{6.051\\(0.9219)} & \makecell{5.704\\(0.9375)} & \makecell{3.658\\(0.9692)} &\makecell{2.027\\(0.9930)}    & 0.9395                     
\end{tabular}
}
\vspace{1 pt}
\caption{\small For a \alg model that has been trained on 10 datasets each having a different boundary condition (Rayleigh number) as $Ra\in [2, 90]\times10^5$ with $Pr = 1$, the super-resolution performance evaluation is reported for: a Rayleigh number within the range of boundary conditions of the training sets (\textit{i.e.}, $Ra = 5\times10^6$), Rayleigh numbers slightly below and above the range of boundary conditions of the training sets (\textit{i.e.}, $Ra = 1\times10^5$ and $Ra = 1\times10^7$ respectively), and Rayleigh numbers far below and above the range of boundary conditions of the training sets (\textit{i.e.}, $Ra = 1\times10^4$ and $Ra = 1\times10^8$ respectively).}
\label{tab:diffBoundary}
\end{table*}

\section{Experiments}
In all the experiments, we use an Adam optimizer with learning rate of $10^{-2}$, and $l_1$ regularization for the loss function, 3000 random samples per epoch, and train for 100 epochs.
\subsection{Prediction Loss vs. Equation Loss}\label{sec:gamma}
In this part, we investigate the influence of the importance given to the Equation loss and the prediction loss (see, Section~\ref{ssec:loss}) on the performance of \alg. As presented in Eqn.\,\ref{eq:loss}, the total loss ($\mathcal{L}$) comprises of the Prediction loss ($\mathcal{L}_p$) and the Equation loss ($\mathcal{L}_e$) where the Equation loss is weighted with a scaling coefficient $\gamma$. Accordingly, we study the influence of the hyperparameter $\gamma$ in the loss function given in Eqn.\,\ref{eq:loss} on the accuracy of \alg. For this purpose, we consider $\gamma\,\in\,\lbrace 0, 0.0125, 0.025, 0.05, 0.1, 0.2, 0.4, 0.8, 1.0 \rbrace$. Considering both Softplus and Swish activation functions in the continuous decoding network, we have found that the results obtained by the latter one outperform the ones obtained by the former one, for which, the values of the evaluation metrics for \alg trained with each of the $\gamma$ values are presented in Table\,\ref{table:gamma}. $\gamma = 0$ indicates a loss function which only depends on the Prediction loss, and the physical aspects (PDE imposed constraints) of the predicted high-resolution data are not accounted for. As presented in Table \ref{table:gamma}, the best performance on the validation set is achieved for a loss function with the Equation loss weighting coefficient of $\gamma = 0.05$. Allover this work we refer to this optimum weighting coefficient as $\gamma^*$ and perform the training tasks using the loss function in Eqn.\,\ref{eq:loss} where the Equation loss is weighted with $\gamma^* = 0.05$. As presented in Table \ref{table:gamma}, a model that is trained with $\gamma = 0$, which only focuses on the data (\textit{i.e.}, uses Prediction loss only) and does not account for the physics and the PDE constraints (\textit{i.e.}, ignores Equation loss) underperforms compared to the trained model with $\gamma = \gamma^*$. On the other hand models trained with a significant focus on the physical constraints only (\textit{i.e.}, large $\gamma$) underperform in super-resolving the low-resolution data. In general, a balance between the focus of the \alg on the model and the physical constraints leads to an optimal super-resolution performance. In achieving that, in Eqn.\,\ref{eq:loss}, the Prediction loss ($\mathcal{L}_p$) captures the global structure of the data and the Equation loss ($\mathcal{L}_e$) further guides the model in accurately reconstructing the local structure of the data.
\begin{figure*}[t!]
    \centering
    \psfrag{t1}{$n_1$}
        \includegraphics[width=\linewidth]{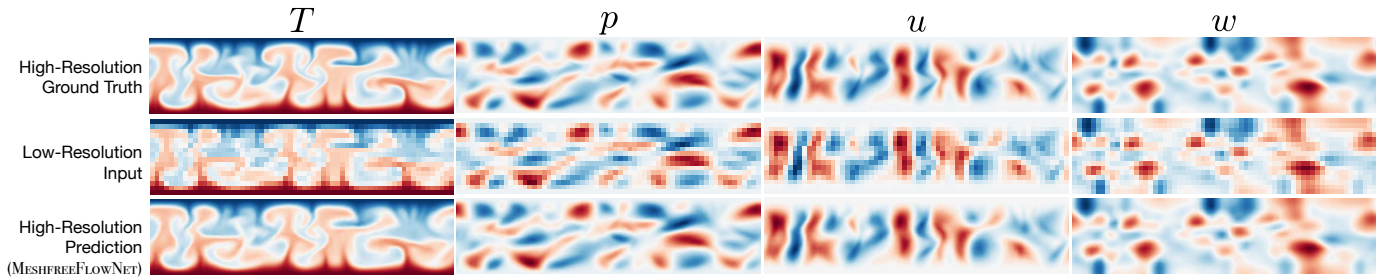}
        \psfrag{n1}{$n_1$}
        \caption{\small Sample tuples of low-resolution input data for \alg, the high-resolution super-resolved data by \alg, and the ground truth high-resolution data for the 4 physical parameters of the Rayleigh–Bénard problem, \textit{i.e.}, $T, p, u, w$ as the temperature, pressure, and the $x$ and $z$ components of the velocity. The contours correspond to the \alg model evaluation presented in Table \ref{tab:diffBoundary} for $Ra$ = $1\times10^8$.}
    \label{fig:reayleighBenard3}
\end{figure*}
\begin{figure*}[t]
    \centering
    \begin{subfigure}[t]{0.3265\textwidth}
        \centering
        \includegraphics[width=.8\linewidth]{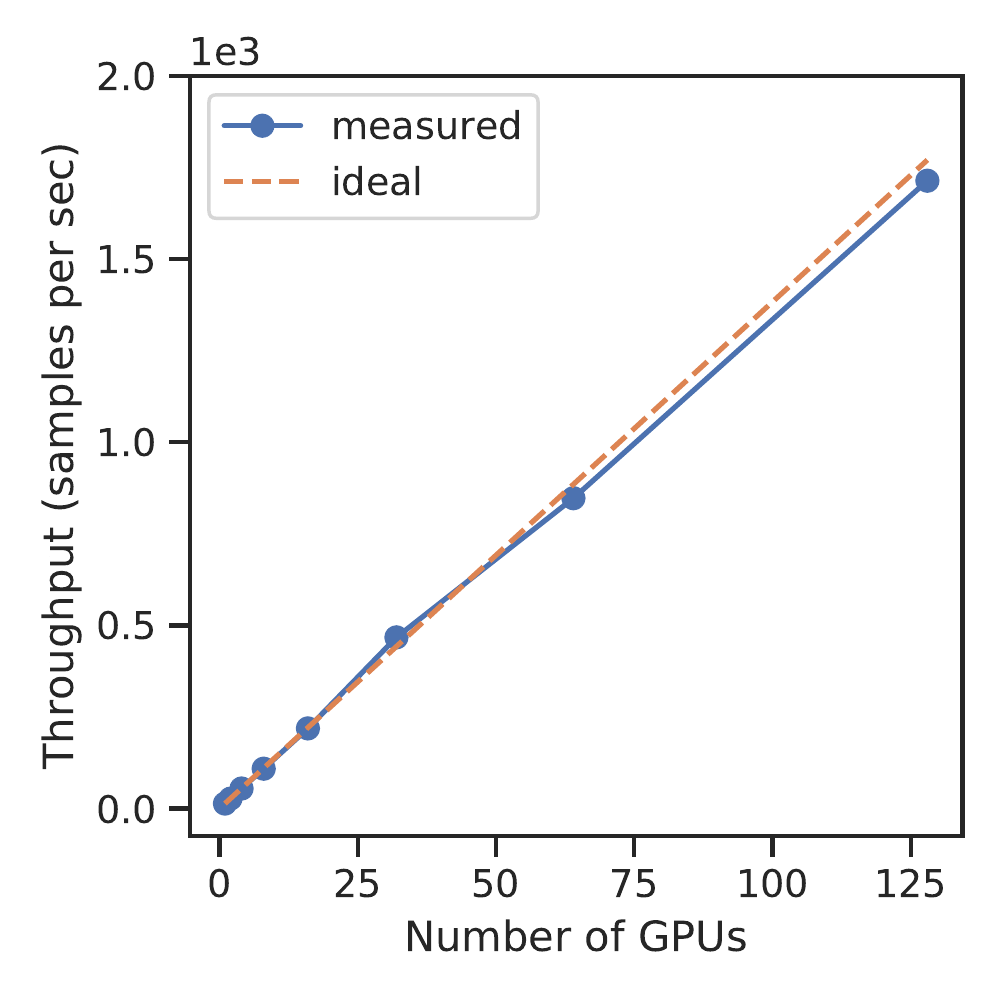}
        \caption{\small Throughput vs. Num. of GPUs}\label{fig:scaling_a}
    \end{subfigure}%
    \begin{subfigure}[t]{0.32\textwidth}
        \centering
        \includegraphics[width=.8\linewidth]{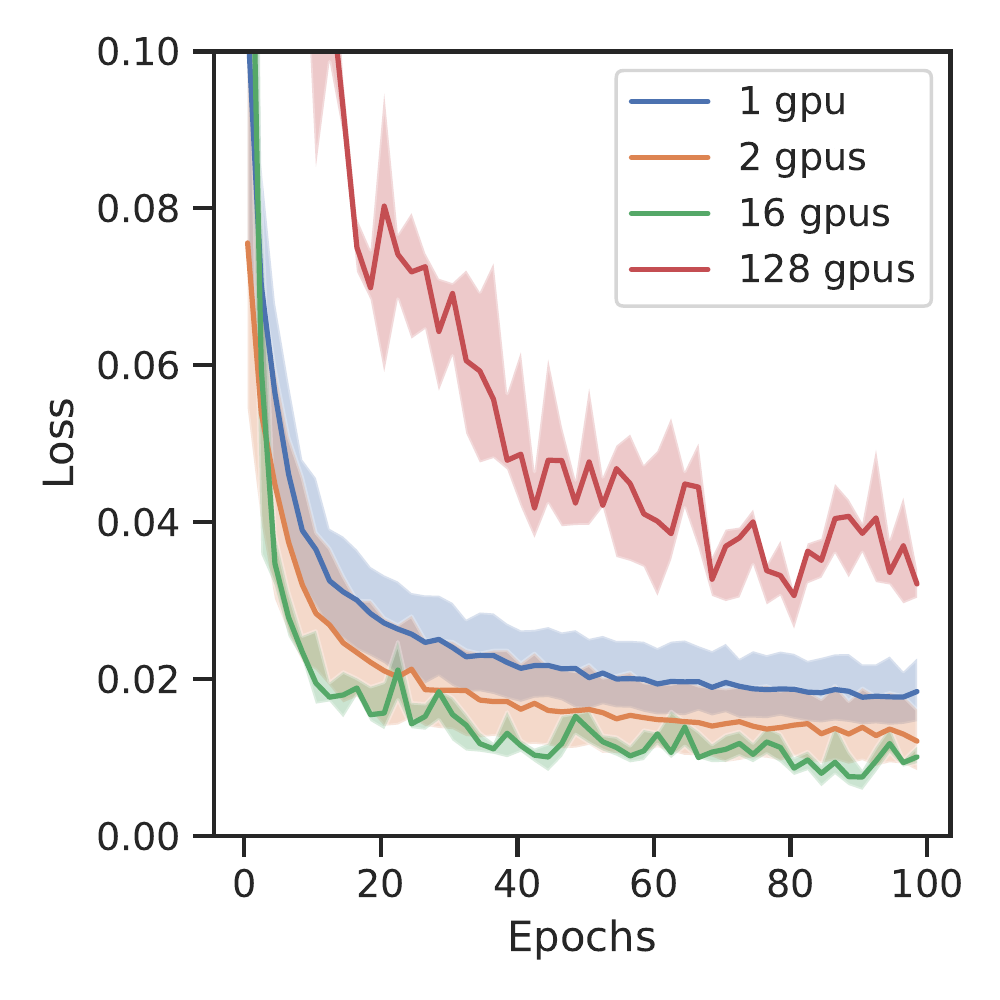}
        \caption{\small Loss vs. Num. of Epochs}\label{fig:scaling_b}
    \end{subfigure}%
    \begin{subfigure}[t]{0.32\textwidth}
        \centering
        \includegraphics[width=.8\linewidth]{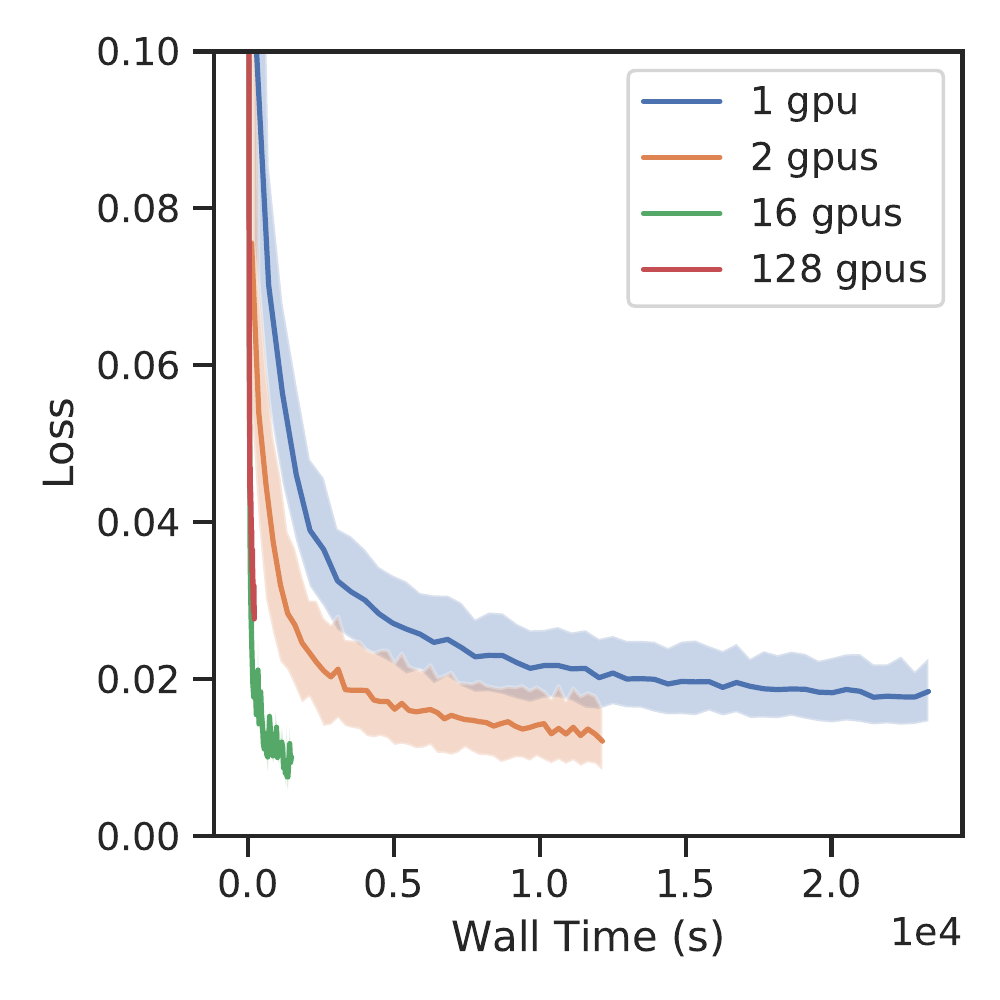}
        \caption{\small Loss vs. Wall Time}\label{fig:scaling_c}
    \end{subfigure}
    \caption{\small Scaling performance on up to 128 GPUs on NERSC Cori GPU cluster. Scaling efficiency is close to ideal performance, at 96\% with all 128 GPUs. Training convergence is greatly accelerated by scaling across more GPUs. The shaded band in Fig.~\ref{fig:scaling_b}-\ref{fig:scaling_c} denotes the $(10\%, 90\%)$ noise interval. \textit{Throughput} refers to the number of samples per second.}\label{fig:scaling}
\end{figure*}

\subsection{\alg vs. Baseline Models}
In this section, we present a comparison between the performance of our proposed \alg framework for super-resolving low-resolution data against two baselines, namely: a classic trilinear interpolation algorithm (Baseline (I)), and a deep learning based 3D U-Net model (Baseline (II)). Specifically for the 3D U-Net model for Baseline (II), we use the same U-Net backbone as in our \alg framework, with the difference being that while \alg uses the 3D U-Net to generate a latent context grid, the baseline U-Net continues with 3D up-sampling and transpose-convolution operations up to the target high-resolution space (see, Fig.~\ref{fig:arch}). Table \ref{tab:baselines} presents the comparison of the \alg with the baseline models. As shown in Table \ref{tab:baselines}, the Baseline (I) is a purely interpolation-based approach and fails to reconstruct the high-resolution data and resolve the fine-scale details, leading to large errors in flow-based evaluation metrics that characterize the flow dynamics. The extremely large normalized mean absolute error (NMAE) of the calculated $rms$ velocity for the fine-scale solution found by Baseline (I) indicates that a merely interpolative scheme cannot accurately reconstruct the fine-scale \textit{local} dynamics (\textit{e.g.}, the flow velocities). Moreover, the NMAE of the total kinetic energy ($E_{tot})$, as a \textit{global} characterizing parameter of the turbulent dynamics, is also very large for the high-resolution solution by Baseline (I). On the other hand, the deep learning based Baseline (II) which utilizes the 3D U-Net model directly maps the low-resolution data to the high-resolution space, achieves better performance compared to Baseline (I). However, as presented in Table \ref{tab:baselines}, our \alg model performs significantly better than Baselines (I) and (II). The specification of $\gamma$ in Table \ref{tab:baselines} refers to the weighting coefficient of the Equation loss component in the loss function in Eqn.\,\ref{eq:loss}. In Table  \ref{tab:baselines}, $\gamma = 0$ indicates that only the prediction loss has been considered for training the \alg model, whereas $\gamma = \gamma^*$ refers to the optimum value for the weighting coefficient of the Equation loss from the ablation study (see, Table \ref{table:gamma}, Sec.~\ref{sec:gamma}).

\subsection{Generalizability of \alg}

We further evaluate the robustness of \alg for resolution enhancement of low-resolution datasets that have physical initial and boundary conditions different from the datasets the \alg model has been trained on. We refer to such initial and boundary conditions as \textit{unseen} initial/boundary conditions. In order to investigate the generalizability of \alg on unseen initial and boundary conditions, we study the effect of each condition separately in the following setups.
\subsubsection{Unseen Physical Initial Conditions}\label{sec:InitialCond}
We investigate the robustness of a trained \alg for enhancing the resolution of a low-resolution unseen data with physical initial conditions different than the training datasets that the \alg has been trained on. Table \ref{table:initialCond} shows the performance of the \alg on a dataset with unseen initial conditions. The first row of Table \ref{table:initialCond} shows the values of the evaluation metrics for unseen test data when \alg is trained only on one dataset whereas the second row shows the values of the evaluation metrics when \alg is trained on 10 datasets each with a different initial condition. As can be observed from the results, the performance of \alg on unseen cases can be improved by training on a more diverse set of initial conditions.

\subsubsection{Unseen Physical Boundary Conditions}
We further investigate the robustness of a trained \alg for super-resolving a low-resolution unseen dataset with physical boundary conditions different from the training datasets that the \alg has been trained on. As the use of more datasets in Sec.~\ref{sec:InitialCond} was shown to be effective in improving the performance of \alg for super-resolution, here we use a dataset that comprises of 10 different sets of boundary conditions (\textit{i.e.} different Rayleigh numbers of $Ra$ $\in [2, 90]\times10^5$, corresponding to Reynolds numbers of up to 10,000). Table \ref{tab:diffBoundary} shows the performance of such a trained \alg for 5 different test datasets each with a different Rayleigh number boundary condition. In Table \ref{tab:diffBoundary}, \alg's performance is evaluated for a Rayleigh number within the range of boundary conditions of the training sets (\textit{i.e.}, $Ra = 5\times10^6$), for Rayleigh numbers slightly below and above the range of boundary conditions of the training sets (\textit{i.e.}, $Ra = 1\times10^5$ and $Ra = 1\times10^7$ respectively), and for Rayleigh numbers far below and above the range of boundary conditions of the training sets (\textit{i.e.}, $Ra = 1\times10^4$ and $Ra = 1\times10^8$ respectively). As presented in Table \ref{tab:diffBoundary}, \alg achieves a good performance not only on the boundary conditions within the range of Rayleigh number boundary conditions it has been trained on, but also on the unseen boundary conditions far out of the range of Rayleigh number boundary conditions it has been trained on. This illustrates the fact that a trained \alg model can generalize well to unseen boundary conditions.

\subsection{Scalability of \alg}
%

Last but not least, we study the scalability of the \alg model to study its applicability to larger problems that require orders-of-magnitude more compute. The scaling results are presented in Fig.~\ref{fig:scaling}. Figs.~\ref{fig:scaling_a} demonstrates that an almost-ideal scaling performance can be achieved for the \alg model on up to 128 GPU workers, achieving approximately $96.80\%$ scaling efficiency. In Fig.~\ref{fig:scaling_b}, we show the convergence for the model training loss with respect to the number of epochs. In Fig.~\ref{fig:scaling_c}, we show the loss convergence with respect to total wall time, where an increasing number of GPU workers leads to a drastic decrease in total training time. As the models achieve similar levels of losses after 100 epochs, yet the training throughput scales almost linearly with the number of GPUs, we see a close to an ideal level of scaling for convergence speed on up to 16 GPUs. A small anomaly regarding the loss curve for 128 GPUs where the loss does not decrease to an ideal level after 100 epochs shows that a very large batch size could lead to diminishing scalability with respect to model convergence. This has been observed in numerous machine learning scaling studies and requires further investigations within the community.
\section{Conclusion and Future Work}
In this work, for the first time, we presented the \alg, a physics-constrained super-resolution framework, that can produce continuous super-resolution outputs, would allow imposing arbitrary combinations of PDE constraints and could be evaluated on arbitrary-sized spatio-temporal domains due to its fully-convolutional nature. We further demonstrated that \alg can recover a wide range of important physical flow quantities (\textit{e.g.}, including Turbulent Kinetic Energy, Kolmogorov Time and Length Scales, etc.) by accurately super-resolving turbulent flows significantly better than traditional (trilinear interpolation) and deep learning based (3D U-Net) baselines. We further illustrated the scalability of \alg to a large cluster of GPU nodes with a high speed interconnect, demonstrating its applicability to problems that require orders of magnitude more computational resources.

Future work includes exploring the applicability of \alg to other physical applications beyond 2D Rayleigh Bernard convection. One interesting direction to pursue is to explore the use of 4D spatio-temporal convolution operators such as those proposed by \citet{choy20194d} to further extend this framework to 4D space-time simulations. That will open up a wide range of applications in Turbulence modeling, where statistical priors between the low-resolution simulation and subgrid-scale physics can be learned from $3+1$D Direct Numerical Simulations (DNS). The scalability of the model on HPC clusters will be critical in learning from such large scale datasets, where single node training will be prohibitively slow. The fully convolutional nature of the \alg framework, along with the demonstrated scalability makes it well poised for such challenges. Moreover, due to the generalizability of our PDE constrained framework, it would be interesting to apply this framework on applications beyond turbulent flows.

\section*{Acknowledgements}
This research used resources of the National Energy Research Scientific Computing Center (NERSC), a DOE Office of Science User Facility supported by the Office of Science of the U.S. Department of Energy under Contract No. DE-AC02-05CH11231. The research was performed at the Lawrence Berkeley National Laboratory for the U.S. Department of Energy under Contract No. DE340AC02- 05CH11231. K. Kashinath is supported by the Intel Big Data Center at NERSC. K. Azizzadenesheli gratefully acknowledges the financial support of Raytheon and Amazon Web Services. A. Anandkumar is supported in part by Bren endowed chair, DARPA PAIHR00111890035 and LwLL grants, Raytheon, Microsoft, Google, and Adobe faculty fellowships. We also acknowledge the Industrial Consortium on Reservoir Simulation Research at Stanford University (SUPRI-B).

\bibliographystyle{unsrtnat}
\small{\bibliography{references}}
\end{document}